\documentclass{article}
\usepackage{arxiv}
\usepackage[utf8]{inputenc}
\usepackage[T1]{fontenc}
\usepackage{url}
\usepackage{booktabs,amsmath,amssymb,amsfonts,microtype,xcolor,caption}
\usepackage{graphicx}
\usepackage[colorlinks,allcolors=blue!55!black]{hyperref}
\newcommand{\sama}{SAMA}

\title{\sama: Training-Free Consensus Ridge Tracking for Camera-Based Heart Rate on Five Public Datasets}
\author{Sudhakar Mishra\\
  Department of Artificial Intelligence\\
  Sardar Vallabhbhai National Institute of Technology (SVNIT)\\
  Surat, India}
\renewcommand{\headeright}{Preprint}
\renewcommand{\undertitle}{Preprint}
\renewcommand{\shorttitle}{SAMA-v2: Training-Free Consensus Ridge Tracking for rPPG}
\hypersetup{pdftitle={SAMA-v2: Training-Free Consensus Ridge Tracking for Camera-Based Heart Rate on Five Public Datasets},
pdfauthor={Sudhakar Mishra},pdfkeywords={remote photoplethysmography, heart rate, training-free, Viterbi tracking}}

\begin{document}
\maketitle

\begin{abstract}
Remote photoplethysmography (rPPG) is usually benchmarked by one heart rate per video, a figure that hides the short windows in which a tele-consultation would actually read the pulse. We present \sama, a training-free estimator that (i) forms a quality-weighted consensus of POS and CHROM chrominance spectra, (ii) discounts frequencies occupied by pulse-free references derived from the same frames (frame motion, skin-mask geometry and background brightness), (iii) follows the heart rate through time with a Viterbi ridge under a random-walk prior, and (iv) returns a track-filtered, amplitude-normalised waveform with a per-window confidence. No parameter is learned and no dataset-specific tuning is used. On 218 recordings from 111 participants in five public datasets (UBFC-rPPG, PURE, SUMS, M3PD and a two-participant subset of MMPD), all processed with the same fixed face box and scored with a re-implementation of the rPPG-Toolbox protocol, \sama\ has the lowest mean absolute error (MAE) averaged over datasets at every protocol level: 4.40 against 5.13\,bpm for POS per video, 4.80 against 5.43 on 30\,s windows and 5.05 against 6.59 on 10\,s windows. Pooled over participants, the bootstrap interval of its paired difference from POS excludes zero on 30\,s and 10\,s windows (Wilcoxon $p=0.11$ and $0.004$) but not per video, and it does not beat POS on SUMS per video or on the MMPD subset. Withholding the 20\% least confident windows lowers the pooled 10\,s error from 4.06 to 2.33\,bpm. We also show that the per-video protocol has a floor of 1.2--1.6\,bpm on datasets where heart rate drifts, and that skin tone cannot be recovered from uncalibrated video, which motivates the instrumented, skin-tone-stratified collection we propose for the Indian population. The method does not surpass the best published neural networks on UBFC-rPPG; we report where it stands against them without re-running them.
\end{abstract}

\keywords{Remote photoplethysmography \and heart rate \and tele-consultation \and signal quality \and Viterbi tracking \and training-free}

\section{Introduction}
Camera-based pulse measurement could let a tele-consultation platform read heart rate from the patient's own webcam. Two obstacles stand between published accuracy and that use. First, accuracy is known to fall for darker skin and under motion, compression and poor light~\cite{nowara,mmpd}, and the public datasets on which methods are tuned contain few South-Asian participants. Second, benchmark tables report one heart rate per recording~\cite{toolbox}; a clinician needs a reading every few seconds and needs to know when not to trust it.

Supervised networks~\cite{deepphys,tscan,physnet,physformer,efficientphys,rhythmformer,physmamba} lead those tables, but they inherit the population and camera of their training set, and their cross-dataset error on mobile video exceeds 9\,bpm~\cite{toolbox,mmpd}. Classical projections such as POS~\cite{pos} and CHROM~\cite{chrom} need no training and remain strong baselines, yet they treat each window in isolation and fail by picking a wrong spectral peak.

\begin{figure}[t]
\centering
\includegraphics[width=\linewidth]{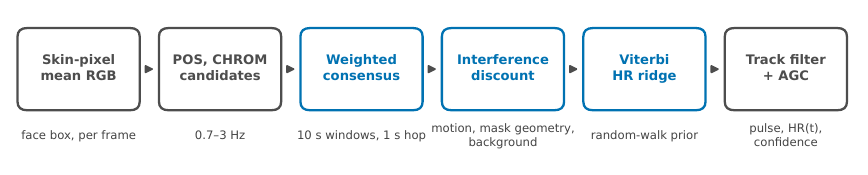}
\caption{\sama. The three blue stages are the additions to a classical chrominance pipeline. Nothing is trained.}
\label{fig:pipe}
\end{figure}

This paper asks how far a method with no learned parameters can go when it uses two facts the classical projections ignore: heart rate is continuous in time, and several artefact sources can be measured from the video itself. Our contributions are:
\begin{itemize}\setlength\itemsep{0pt}\setlength\parskip{1pt}
\item \sama, a training-free pipeline (Fig.~\ref{fig:pipe}) that combines chrominance consensus, an interference discount built from pulse-free references, and Viterbi ridge tracking, and that outputs a waveform, a heart-rate track and a confidence.
\item A like-for-like benchmark of \sama\ and six classical methods on five public datasets from identical colour traces, at three protocol levels, with participant-cluster bootstrap intervals and paired tests.
\item Two measurement findings: a quantified floor of the per-video FFT protocol, and evidence that apparent skin tone from uncalibrated video is not usable for stratification.
\item An honest account of what did not work: sub-face regions and a learned spectral-evidence model.
\end{itemize}

\section{Related work}
\paragraph{Classical methods.} GREEN~\cite{green}, ICA~\cite{ica}, CHROM~\cite{chrom}, PBV~\cite{pbv}, POS~\cite{pos} and LGI~\cite{lgi} map a spatially averaged RGB trace to a pulse signal by a fixed or data-derived projection. The rPPG-Toolbox~\cite{toolbox} provides reference implementations and a common scoring protocol, which we follow.
\paragraph{Neural methods.} DeepPhys~\cite{deepphys}, TS-CAN~\cite{tscan}, PhysNet~\cite{physnet}, PhysFormer~\cite{physformer}, EfficientPhys~\cite{efficientphys}, RhythmFormer~\cite{rhythmformer} and PhysMamba~\cite{physmamba} learn the mapping from video. Published per-video MAE is below 1\,bpm within UBFC-rPPG and PURE and between 0.9 and 17\,bpm across datasets~\cite{toolbox,rhythmformer,physmamba}.
\paragraph{Temporal tracking and signal quality.} Tracking a spectral ridge is standard for wearable PPG under motion; in rPPG, region weighting by signal quality~\cite{distanceppg} is the closest idea. \sama\ differs in using references that cannot contain the pulse to discount artefact frequencies, and in reporting a confidence that supports abstention.

\section{Method}
\label{sec:method}
Let $\mathbf{c}(t)\in\mathbb{R}^3$ be the mean colour of the skin pixels of the face box in frame $t$ (frame rate $f_s$).

\paragraph{Candidates.} Two pulse candidates are computed from $\mathbf{c}$: POS~\cite{pos} (1.6\,s overlap-add) and CHROM~\cite{chrom}. Each is band-passed to 0.7--3\,Hz (42--180\,bpm) and scaled to unit variance, giving $x_k(t)$, $k\in\{1,2\}$.

\paragraph{Consensus spectrogram.} In Hann windows of 10\,s with a 1\,s hop we evaluate the power of each candidate on a 0.5\,bpm grid and normalise it to unit sum, $P_k(w,f)$. The quality of candidate $k$ in window $w$ is the share of its power within $\pm3$\,bpm of its own peak plus half the share around twice that frequency, $q_k(w)$. The consensus is
\begin{equation}
S(w,f)=\sum_k \frac{q_k(w)^{\gamma}}{\sum_j q_j(w)^{\gamma}}\,P_k(w,f),\qquad \gamma=2,
\end{equation}
so a candidate that is spectrally diffuse in a window contributes little there. A frequency whose first harmonic is also present is raised: $S \leftarrow S + \tfrac12\, S(w,2f)\,S(w,f)/\max_f S$.

\paragraph{Interference discount.} Blood volume changes the colour of skin but not which pixels are skin. We therefore take as pulse-free references (a) the mean absolute frame difference, (b) the number of skin pixels, (c) the horizontal and vertical centroid of the skin mask and (d) the brightness of box cells that contain no skin. Each is band-passed and its normalised spectrum $I_r(w,f)$ computed as above. With $I=\max_r I_r$,
\begin{equation}
\tilde S(w,f)=\frac{S(w,f)}{1+\beta\, I(w,f)/\max_f I(w,f)},\qquad \beta=1 .
\end{equation}

\paragraph{Ridge.} Heart rate is modelled as a random walk with standard deviation $\sigma=2.5$\,bpm per second. The track $\hat f(w)$ maximises $\sum_w \log\tilde S(w,f_w) - \sum_w (f_w-f_{w-1})^2/2\sigma^2$, solved exactly by the Viterbi algorithm. The confidence of a window is the share of $\tilde S$ within $\pm3$\,bpm of the track.

\paragraph{Waveform.} Candidates are summed with weights given by their mean on-track power share, and the sum is filtered in the short-time Fourier domain with a flat-topped pass band of about $\pm6.5$\,bpm around the track and its first harmonic. Finally the waveform is divided by the 2\,s moving average of its analytic envelope (AGC), so that each beat has equal weight in a later spectrum.

All constants ($\gamma$, $\beta$, $\sigma$, window, pass band) were set once from first principles before the full benchmark was run and are the same for every dataset. Section~\ref{sec:abl} reports what each stage contributes; the choice to use the face-wide region only was made after seeing that sub-face regions were worse (a limitation discussed in Sec.~\ref{sec:lim}). The method is linear algebra and FFTs and runs in real time on a CPU.

\section{Experimental protocol}
\paragraph{Datasets.} Table~\ref{tab:data} lists the data. UBFC-rPPG~\cite{ubfc} and PURE~\cite{pure} are webcam and machine-vision recordings; SUMS~\cite{sums} was recorded at high altitude during exercise recovery; M3PD~\cite{m3pd} recordings here are $\sim$30\,s handheld front-camera clips of cardiovascular patients; MMPD~\cite{mmpd} is mobile video under varied light and motion. Only two MMPD participants (31 one-minute recordings, Fitzpatrick types III and IV) were available on disk, so MMPD results describe that subset and are not comparable with published full-dataset numbers. One PURE recording (04-02) has a truncated image file and was excluded for all methods.

\begin{table}[t]
\centering\small
\begin{tabular}{@{}lrrrr@{}}
\toprule
Dataset & Rec. & Part. & Minutes & 10\,s win.\\
\midrule
UBFC-rPPG~\cite{ubfc} & 42 & 42 & 46 & 490\\
PURE~\cite{pure} & 58 & 10 & 67 & 733\\
SUMS~\cite{sums} & 40 & 10 & 97 & 1096\\
M3PD~\cite{m3pd} & 47 & 47 & 27 & 248\\
MMPD$^\dagger$~\cite{mmpd} & 31 & 2 & 31 & 341\\
\midrule
Total & 218 & 111 & 269 & 2908\\
\bottomrule
\end{tabular}
\caption{Recordings evaluated. $^\dagger$Two-participant subset.}
\label{tab:data}
\end{table}

\paragraph{Face region.} No face detector was available on the machine that holds the videos, so every recording uses the same fixed box (central 40\% of the width and 60\% of the height) and a colour-threshold skin mask inside it. This is a harder input than the detector-based crop of the toolbox; it is identical for all methods compared here. For each frame we store the skin-pixel mean colour and count on a $16\times16$ grid; all methods start from these traces.

\paragraph{Baselines.} GREEN, ICA, CHROM, LGI, PBV and POS, re-implemented step for step from the rPPG-Toolbox~\cite{toolbox} (ICA uses FastICA instead of JADE) and run on the same skin-pixel mean trace as \sama.

\paragraph{Scoring.} We re-implement \texttt{calculate\_metric\_per\_video} of the toolbox: prediction and contact-PPG reference are detrended ($\lambda=100$), band-passed (first-order Butterworth, 0.75--2.5\,Hz, as in~\cite{toolbox}) and reduced to the periodogram peak. \emph{Per video} applies this to the whole recording. \emph{30\,s} and \emph{10\,s} apply it to windows with 50\% overlap. We report MAE, RMSE, Pearson $r$ and the toolbox SNR. The current toolbox release uses 0.6--3.3\,Hz; Sec.~\ref{sec:sens} repeats the per-video analysis with that band.

\paragraph{Statistics.} Confidence intervals are 95\% percentile bootstrap intervals (10\,000 resamples) of the MAE, resampling participants, not windows. Paired comparisons use the participant-mean difference in absolute error: a two-sided Wilcoxon signed-rank test, Holm-corrected over the six baselines within each dataset and level, and a cluster-bootstrap interval of the difference. For the MMPD subset the resampling unit is the recording, which understates uncertainty. No recording, window or participant was removed on the basis of its error.

\section{Results}
\subsection{Comparison with classical methods}
\begin{figure}[t]
\centering
\includegraphics[width=\linewidth]{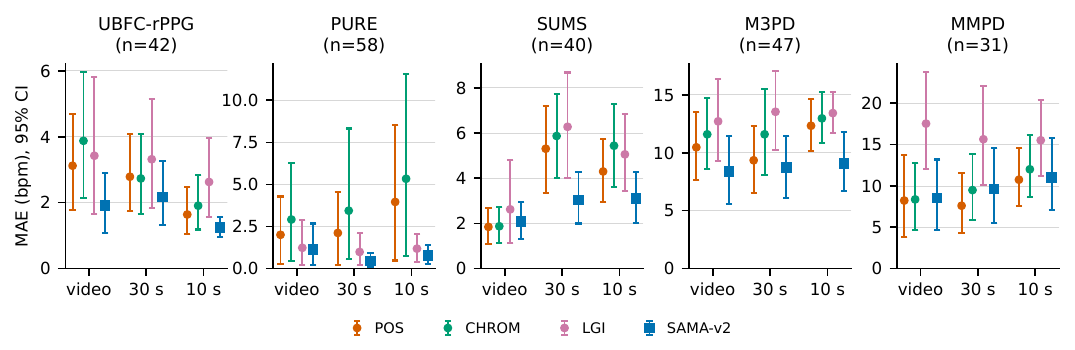}
\caption{MAE with 95\% participant-bootstrap intervals for \sama\ and the three strongest baselines.}
\label{fig:mae}
\end{figure}

Table~\ref{tab:main} and Fig.~\ref{fig:mae} give the main result. Averaged over datasets, \sama\ has the lowest MAE at all three levels, and its advantage over POS, the best baseline, is largest on the shortest windows: $-0.73$\,bpm per video, $-0.63$ on 30\,s and $-1.55$ on 10\,s windows. It is the best method on UBFC-rPPG, PURE and M3PD at every level and on SUMS for windows. It is not the best on SUMS per video (2.08 against 1.85 for POS) or on the MMPD subset at any level (8.51, 9.62 and 11.03 against 8.22, 7.60 and 10.74 for POS).

\begin{table}[t]
\centering\small
\resizebox{\linewidth}{!}{\begin{tabular}{@{}l@{\hspace{5pt}}c@{\hspace{5pt}}c@{\hspace{5pt}}c@{\hspace{5pt}}c@{\hspace{5pt}}c@{\hspace{5pt}}c@{\hspace{4pt}}c@{}}
\toprule
Method & UBFC-rPPG & PURE & SUMS & M3PD & MMPD$^\dagger$ & Mean & Rank\\
\midrule
\multicolumn{8}{@{}l}{\emph{Per video (toolbox protocol)}}\\
GREEN & 7.21 {\scriptsize[3.2, 12.3]} & 7.64 {\scriptsize[0.6, 18.6]} & 8.73 {\scriptsize[4.7, 13.1]} & 17.53 {\scriptsize[14.1, 21.1]} & 19.48 {\scriptsize[13.5, 26.1]} & 12.12 & 6.8 \\
ICA & 4.90 {\scriptsize[1.9, 9.1]} & 2.34 {\scriptsize[0.3, 5.1]} & 5.27 {\scriptsize[2.2, 8.8]} & 13.32 {\scriptsize[9.5, 17.5]} & 19.62 {\scriptsize[13.9, 25.8]} & 9.09 & 5.4 \\
CHROM & 3.88 {\scriptsize[2.1, 6.0]} & 2.92 {\scriptsize[0.5, 6.3]} & 1.87 {\scriptsize[1.1, 2.7]} & 11.59 {\scriptsize[8.6, 14.7]} & 8.36 {\scriptsize[4.6, 12.8]} & 5.72 & 3.2 \\
LGI & 3.42 {\scriptsize[1.7, 5.8]} & 1.23 {\scriptsize[0.2, 2.9]} & 2.62 {\scriptsize[1.1, 4.8]} & 12.71 {\scriptsize[9.3, 16.4]} & 17.52 {\scriptsize[12.0, 23.7]} & 7.50 & 3.6 \\
PBV & 7.11 {\scriptsize[3.6, 11.4]} & 5.10 {\scriptsize[0.5, 12.4]} & 5.85 {\scriptsize[2.1, 10.5]} & 13.00 {\scriptsize[9.4, 16.9]} & 16.02 {\scriptsize[10.0, 22.7]} & 9.42 & 5.4 \\
POS & 3.12 {\scriptsize[1.8, 4.7]} & 2.01 {\scriptsize[0.3, 4.3]} & \textbf{1.85} {\scriptsize[1.1, 2.7]} & 10.46 {\scriptsize[7.6, 13.5]} & \textbf{8.22} {\scriptsize[3.8, 13.8]} & 5.13 & 1.8 \\
\textbf{SAMA-v2} & \textbf{1.92} {\scriptsize[1.1, 2.9]} & \textbf{1.14} {\scriptsize[0.2, 2.7]} & 2.08 {\scriptsize[1.3, 2.9]} & \textbf{8.37} {\scriptsize[5.6, 11.4]} & 8.51 {\scriptsize[4.6, 13.2]} & 4.40 & 1.8 \\
\midrule
\multicolumn{8}{@{}l}{\emph{30\,s windows, 15\,s step}}\\
GREEN & 7.49 {\scriptsize[4.1, 11.9]} & 6.92 {\scriptsize[0.5, 17.8]} & 11.61 {\scriptsize[7.6, 15.8]} & 17.49 {\scriptsize[13.9, 21.2]} & 17.98 {\scriptsize[12.6, 23.8]} & 12.30 & 6.8 \\
ICA & 4.03 {\scriptsize[1.9, 7.0]} & 1.91 {\scriptsize[0.3, 4.3]} & 7.99 {\scriptsize[4.7, 11.5]} & 12.86 {\scriptsize[9.3, 16.8]} & 18.86 {\scriptsize[14.0, 24.2]} & 9.13 & 4.8 \\
CHROM & 2.73 {\scriptsize[1.7, 4.1]} & 3.44 {\scriptsize[0.6, 8.3]} & 5.87 {\scriptsize[4.0, 7.7]} & 11.59 {\scriptsize[8.1, 15.5]} & 9.49 {\scriptsize[5.9, 13.8]} & 6.62 & 3.0 \\
LGI & 3.32 {\scriptsize[1.8, 5.2]} & 0.99 {\scriptsize[0.2, 2.1]} & 6.27 {\scriptsize[4.0, 8.7]} & 13.53 {\scriptsize[10.2, 17.1]} & 15.63 {\scriptsize[10.1, 22.1]} & 7.95 & 3.8 \\
PBV & 6.13 {\scriptsize[3.5, 9.2]} & 4.79 {\scriptsize[0.3, 13.1]} & 10.17 {\scriptsize[5.9, 15.3]} & 14.54 {\scriptsize[10.9, 18.5]} & 17.14 {\scriptsize[11.8, 23.2]} & 10.55 & 5.8 \\
POS & 2.79 {\scriptsize[1.7, 4.1]} & 2.12 {\scriptsize[0.2, 4.6]} & 5.30 {\scriptsize[3.3, 7.2]} & 9.34 {\scriptsize[6.5, 12.3]} & \textbf{7.60} {\scriptsize[4.3, 11.6]} & 5.43 & 2.4 \\
\textbf{SAMA-v2} & \textbf{2.18} {\scriptsize[1.3, 3.3]} & \textbf{0.45} {\scriptsize[0.1, 0.9]} & \textbf{3.04} {\scriptsize[2.0, 4.3]} & \textbf{8.71} {\scriptsize[6.1, 11.4]} & 9.62 {\scriptsize[5.5, 14.6]} & 4.80 & 1.4 \\
\midrule
\multicolumn{8}{@{}l}{\emph{10\,s windows, 5\,s step}}\\
GREEN & 6.88 {\scriptsize[4.1, 10.4]} & 6.68 {\scriptsize[0.8, 16.4]} & 10.07 {\scriptsize[6.7, 13.7]} & 16.63 {\scriptsize[14.1, 19.3]} & 18.83 {\scriptsize[14.2, 23.9]} & 11.82 & 6.8 \\
ICA & 2.64 {\scriptsize[1.3, 4.7]} & 1.89 {\scriptsize[0.4, 4.0]} & 7.22 {\scriptsize[4.7, 9.8]} & 12.59 {\scriptsize[10.3, 15.1]} & 19.39 {\scriptsize[15.3, 24.0]} & 8.75 & 4.6 \\
CHROM & 1.91 {\scriptsize[1.2, 2.8]} & 5.34 {\scriptsize[0.8, 11.6]} & 5.43 {\scriptsize[3.6, 7.3]} & 12.96 {\scriptsize[10.9, 15.2]} & 12.00 {\scriptsize[8.6, 16.2]} & 7.53 & 4.0 \\
LGI & 2.63 {\scriptsize[1.6, 4.0]} & 1.18 {\scriptsize[0.4, 2.0]} & 5.05 {\scriptsize[3.4, 6.8]} & 13.41 {\scriptsize[11.7, 15.3]} & 15.51 {\scriptsize[11.2, 20.4]} & 7.56 & 3.8 \\
PBV & 5.14 {\scriptsize[3.3, 7.3]} & 5.47 {\scriptsize[0.7, 13.7]} & 9.41 {\scriptsize[5.8, 13.6]} & 12.73 {\scriptsize[10.4, 15.2]} & 17.69 {\scriptsize[13.0, 22.8]} & 10.09 & 5.4 \\
POS & 1.64 {\scriptsize[1.0, 2.5]} & 3.96 {\scriptsize[0.5, 8.5]} & 4.30 {\scriptsize[2.9, 5.7]} & 12.32 {\scriptsize[10.2, 14.6]} & \textbf{10.74} {\scriptsize[7.6, 14.5]} & 6.59 & 2.2 \\
\textbf{SAMA-v2} & \textbf{1.25} {\scriptsize[1.0, 1.6]} & \textbf{0.78} {\scriptsize[0.3, 1.4]} & \textbf{3.10} {\scriptsize[2.0, 4.3]} & \textbf{9.07} {\scriptsize[6.7, 11.8]} & 11.03 {\scriptsize[7.1, 15.8]} & 5.05 & 1.2 \\
\bottomrule
\end{tabular}
}
\caption{Mean absolute error in bpm [95\% participant-bootstrap interval]. Mean: unweighted mean over the five datasets. Rank: mean rank over datasets (1 = best). Bold: lowest error in the column and block. $^\dagger$Two-participant subset; intervals resample recordings.}
\label{tab:main}
\end{table}

Table~\ref{tab:paired} reports paired differences. Pooled over all 111 participants, \sama\ is better than POS by 1.23\,bpm on 30\,s windows (95\% CI 0.05 to 2.45; Wilcoxon $p=0.11$) and by 1.56\,bpm on 10\,s windows (0.53 to 2.85; Wilcoxon $p=0.004$); per video the pooled difference of 0.83\,bpm has an interval that includes zero ($-1.87$ to $0.05$). Against CHROM and LGI the pooled difference excludes zero at every level. Within single datasets most differences from POS do not survive Holm correction: the datasets have 10 to 47 participants and the intervals in Table~\ref{tab:main} overlap. We therefore claim an improvement over classical methods in aggregate and for short windows, not on every dataset.

\begin{table}[t]
\centering\footnotesize
\resizebox{\linewidth}{!}{\begin{tabular}{@{}l@{\hspace{4pt}}l@{\hspace{6pt}}c@{\hspace{6pt}}c@{\hspace{6pt}}c@{\hspace{6pt}}c@{\hspace{6pt}}c@{\hspace{6pt}}c@{}}
\toprule
Level & vs. & UBFC & PURE & SUMS & M3PD & MMPD$^\dagger$ & Pooled [95\% CI]\\
\midrule
video & POS & $-1.20$ & $-0.86$$^{*}$ & $+0.24$ & $-2.09$ & $+0.28$ & $-0.83$ {\scriptsize[-1.87, 0.05]} \\
 & CHROM & $-1.96$$^{*}$ & $-1.77$ & $+0.21$ & $-3.21$ & $+0.14$ & $-1.48$$^{*}$ {\scriptsize[-2.64, -0.49]} \\
 & LGI & $-1.50$ & $-0.09$ & $-0.54$ & $-4.33$$^{*}$ & $-9.02$$^{*}$$^{\ddagger}$ & $-2.63$$^{*}$ {\scriptsize[-4.40, -1.00]} \\
\midrule
30 s & POS & $-0.61$$^{*}$ & $-1.67$$^{*}$$^{\ddagger}$ & $-2.26$$^{*}$ & $-0.64$ & $+2.02$ & $-1.23$$^{*}$ {\scriptsize[-2.45, -0.05]} \\
 & CHROM & $-0.56$ & $-2.99$$^{*}$$^{\ddagger}$ & $-2.82$$^{*}$$^{\ddagger}$ & $-2.88$ & $+0.13$ & $-2.14$$^{*}$ {\scriptsize[-3.61, -0.94]} \\
 & LGI & $-1.14$$^{*}$ & $-0.54$ & $-3.23$$^{*}$$^{\ddagger}$ & $-4.82$$^{*}$$^{\ddagger}$ & $-6.01$$^{*}$$^{\ddagger}$ & $-2.66$$^{*}$ {\scriptsize[-3.91, -1.41]} \\
\midrule
10 s & POS & $-0.39$ & $-3.18$$^{*}$ & $-1.20$$^{*}$ & $-3.26$$^{*}$$^{\ddagger}$ & $+0.29$ & $-1.56$$^{*}$ {\scriptsize[-2.85, -0.53]} \\
 & CHROM & $-0.66$$^{*}$ & $-4.56$$^{*}$ & $-2.34$$^{*}$$^{\ddagger}$ & $-3.90$$^{*}$$^{\ddagger}$ & $-0.97$ & $-2.59$$^{*}$ {\scriptsize[-4.26, -1.33]} \\
 & LGI & $-1.38$$^{*}$ & $-0.40$$^{*}$ & $-1.95$$^{*}$$^{\ddagger}$ & $-4.35$$^{*}$$^{\ddagger}$ & $-4.47$$^{*}$$^{\ddagger}$ & $-1.97$$^{*}$ {\scriptsize[-2.73, -1.21]} \\
\bottomrule
\end{tabular}
}
\caption{Paired difference in MAE, \sama\ minus baseline (bpm; negative favours \sama). $^{*}$bootstrap 95\% interval excludes zero; $^{\ddagger}$Holm-adjusted Wilcoxon $p<0.05$. $^\dagger$Recording-level resampling.}
\label{tab:paired}
\end{table}

\begin{figure}[t]
\centering
\includegraphics[width=\linewidth]{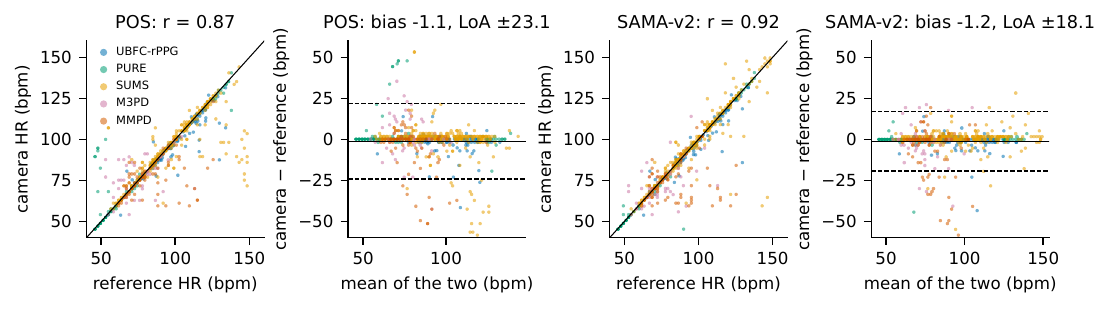}
\caption{Agreement with contact PPG on all 765 thirty-second windows. Solid: bias; dashed: 95\% limits of agreement.}
\label{fig:ba}
\end{figure}

Table~\ref{tab:second} gives RMSE, correlation and SNR per video. On pooled 30\,s windows (Fig.~\ref{fig:ba}) the limits of agreement narrow from $\pm23.1$\,bpm for POS to $\pm18.1$\,bpm and the correlation rises from 0.87 to 0.92; the bias is $-1.2$\,bpm for both.

\begin{table}[t]
\centering\footnotesize
\resizebox{\linewidth}{!}{\begin{tabular}{@{}l*{5}{@{\hspace{7pt}}c@{\hspace{3pt}}c@{\hspace{3pt}}c}@{}}
\toprule
 & \multicolumn{3}{c}{UBFC-rPPG} & \multicolumn{3}{c}{PURE} & \multicolumn{3}{c}{SUMS} & \multicolumn{3}{c}{M3PD} & \multicolumn{3}{c}{MMPD$^\dagger$}\\
Method & RMSE & $r$ & SNR & RMSE & $r$ & SNR & RMSE & $r$ & SNR & RMSE & $r$ & SNR & RMSE & $r$ & SNR\\
\midrule
GREEN & 16.66 & 0.65 & -5.8 & 21.75 & 0.47 & 4.0 & 15.51 & 0.56 & -8.5 & 21.43 & 0.08 & -12.9 & 26.38 & -0.07 & -12.7 \\
ICA & 13.08 & 0.80 & 0.3 & 11.01 & 0.89 & 12.4 & 10.81 & 0.78 & -4.2 & 19.45 & -0.12 & -10.3 & 25.82 & -0.16 & -13.3 \\
CHROM & 7.49 & 0.92 & 0.4 & 10.76 & 0.90 & 8.3 & 3.10 & 0.98 & -0.5 & 15.95 & 0.20 & -10.4 & 14.53 & 0.56 & -8.9 \\
LGI & 7.79 & 0.92 & 0.6 & 6.02 & 0.97 & 12.3 & 6.39 & 0.92 & -1.5 & 17.71 & 0.16 & -11.8 & 24.14 & -0.02 & -11.6 \\
PBV & 14.83 & 0.76 & -3.6 & 15.40 & 0.78 & 7.2 & 12.04 & 0.72 & -6.6 & 18.49 & 0.24 & -11.2 & 24.16 & -0.03 & -12.2 \\
POS & 5.84 & 0.95 & 2.6 & 8.60 & 0.93 & 10.3 & 3.13 & 0.98 & -0.8 & 14.77 & 0.32 & -10.0 & 16.49 & 0.24 & -7.9 \\
\textbf{SAMA-v2} & 3.58 & 0.98 & 8.5 & 6.02 & 0.97 & 21.2 & 3.37 & 0.98 & 8.6 & 13.27 & 0.49 & -3.6 & 15.06 & 0.43 & -2.4 \\
\bottomrule
\end{tabular}
}
\caption{Per-video RMSE (bpm), Pearson $r$ and SNR (dB, toolbox definition).}
\label{tab:second}
\end{table}

\subsection{Position relative to published neural results}
We did not run neural networks (no deep-learning framework was installed on the data machine), so this subsection compares with published numbers that were obtained with a detector-based face crop and, for MMPD and M3PD, different recordings or splits. It is context, not a controlled comparison.

On \textbf{PURE}, \sama\ reaches 1.14\,bpm per video (0.48 with the 0.6--3.3\,Hz band). Published cross-dataset networks trained on UBFC-rPPG report 3.69 to 12.92\,bpm in the toolbox benchmark~\cite{toolbox}, 0.97 for RhythmFormer~\cite{rhythmformer} and 1.20 for PhysMamba~\cite{physmamba}; within-dataset training reaches about 0.2--0.3~\cite{physmamba}. On \textbf{UBFC-rPPG}, \sama\ reaches 1.92\,bpm; networks trained on PURE report 0.89~\cite{rhythmformer} to 2.07~\cite{toolbox}, so the best of them are clearly ahead. On \textbf{M3PD}, the dataset paper reports 15.5 to 28.2\,bpm for face-only networks on the clinical cohort and 7.4 for its dual-view face-plus-fingertip model~\cite{m3pd}; \sama\ obtains 8.37\,bpm from the face alone on 47 clinical recordings. On \textbf{SUMS}, the dataset paper reports below 0.5\,bpm for supervised cross-subject training~\cite{sums}, well below our 2.08. The toolbox's own numbers for classical methods (POS 3.67 on PURE, 4.00 on UBFC-rPPG) are higher than ours for the same algorithms (2.01 and 3.12), which shows that skin-pixel averaging matters as much as the projection.

\subsection{Ablation}
\label{sec:abl}
\begin{figure}[t]
\centering
\includegraphics[width=\linewidth]{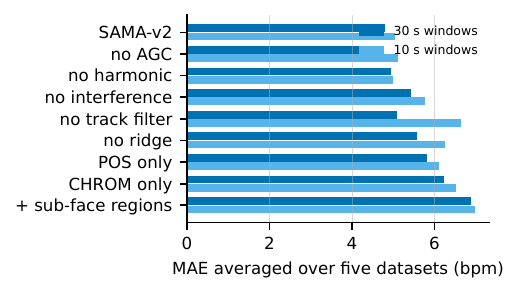}
\caption{Five-dataset mean MAE of each ablation.}
\label{fig:abl}
\end{figure}

Table~\ref{tab:abl} and Fig.~\ref{fig:abl} remove one stage at a time. On windows, the ridge, the interference discount and the use of both projections each reduce the five-dataset mean by 0.6 to 1.5\,bpm. The track filter matters most on PURE. AGC is the one stage whose effect is mixed: it lowers the per-video error on UBFC-rPPG from 2.56 to 1.92 but raises it per video on PURE, SUMS and the MMPD subset, and on M3PD 30\,s windows; the five-dataset per-video mean is unchanged (4.40 against 4.41). Adding $2\times2$ and $4\times4$ sub-face regions to the consensus makes the mean error about 2\,bpm worse; with a fixed box and a threshold skin mask, small regions carry coherent mask-edge artefacts that the quality weight mistakes for a pulse.

\begin{table}[t]
\centering\footnotesize
\resizebox{\linewidth}{!}{\begin{tabular}{@{}l@{\hspace{5pt}}c@{\hspace{5pt}}c@{\hspace{5pt}}c@{\hspace{5pt}}c@{\hspace{5pt}}c@{\hspace{7pt}}c@{\hspace{5pt}}c@{\hspace{5pt}}c@{}}
\toprule
 & \multicolumn{5}{c}{30\,s windows} & \multicolumn{3}{c}{Mean of five}\\
Variant & UBFC & PURE & SUMS & M3PD & MMPD$^\dagger$ & 30\,s & 10\,s & video\\
\midrule
\textbf{SAMA-v2} & 2.18 & 0.45 & 3.04 & 8.71 & 9.62 & 4.80 & 5.05 & 4.40 \\
no AGC & 2.18 & 0.51 & 3.18 & 7.66 & 9.20 & 4.55 & 5.11 & 4.41 \\
no harmonic & 2.16 & 0.45 & 2.74 & 9.53 & 9.79 & 4.93 & 4.99 & 4.57 \\
no interference & 2.19 & 0.48 & 3.70 & 9.45 & 11.28 & 5.42 & 5.76 & 5.01 \\
no track filter & 2.03 & 1.30 & 5.20 & 8.86 & 8.11 & 5.10 & 6.64 & 4.87 \\
no ridge & 2.29 & 0.74 & 4.82 & 10.05 & 10.00 & 5.58 & 6.24 & 4.92 \\
POS only & 2.36 & 0.54 & 2.40 & 11.14 & 12.61 & 5.81 & 6.10 & 5.25 \\
CHROM only & 2.08 & 1.92 & 3.50 & 12.41 & 11.23 & 6.23 & 6.51 & 5.33 \\
+ sub-face regions & 2.07 & 0.43 & 6.46 & 9.72 & 15.73 & 6.88 & 6.97 & 6.24 \\
\bottomrule
\end{tabular}
}
\caption{Ablation: MAE (bpm).}
\label{tab:abl}
\end{table}

\paragraph{A learned variant that did not help.} We also trained a gradient-boosted model to score each (window, frequency) cell from 35 spectral, chromatic-signature, spatial-agreement and interference features, leave-one-dataset-out, and decoded its output with the same ridge. Its per-video MAE was 2.09 (UBFC-rPPG), 1.18 (PURE), 3.18 (SUMS), 11.08 (M3PD) and 11.91 (MMPD subset): no better than the training-free method on any dataset but UBFC-rPPG windows. With 111 participants the training set is too small and too heterogeneous for the learned evidence to transfer.

\subsection{Confidence and abstention}
\begin{figure}[t]
\centering
\includegraphics[width=\linewidth]{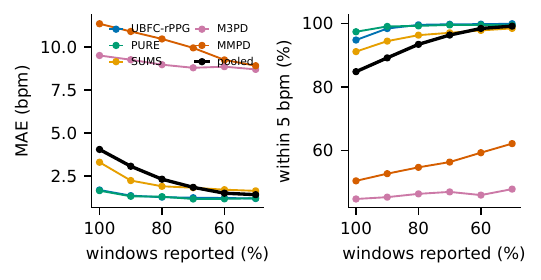}
\caption{Selective reporting by confidence (10\,s windows).}
\label{fig:sel}
\end{figure}

The confidence is informative: over 14\,083 ten-second windows (1\,s hop) its Spearman correlation with absolute error is $-0.40$. Reporting only the most confident windows (Fig.~\ref{fig:sel}) lowers the pooled MAE from 4.06\,bpm to 2.33 at 80\% coverage and 1.43 at 50\%, and raises the share of windows within 5\,bpm from 84.8\% to 93.4\% and 99.2\%. These curves rank windows across datasets with one threshold and were not used to select anything.

\subsection{The per-video protocol has a floor}
\label{sec:sens}
\begin{figure}[t]
\centering
\includegraphics[width=\linewidth]{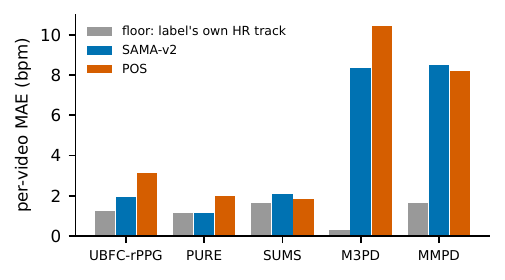}
\caption{Per-video MAE of an oracle built from the reference's own heart-rate track, against \sama\ and POS.}
\label{fig:floor}
\end{figure}

\begin{figure}[t]
\centering
\includegraphics[width=\linewidth]{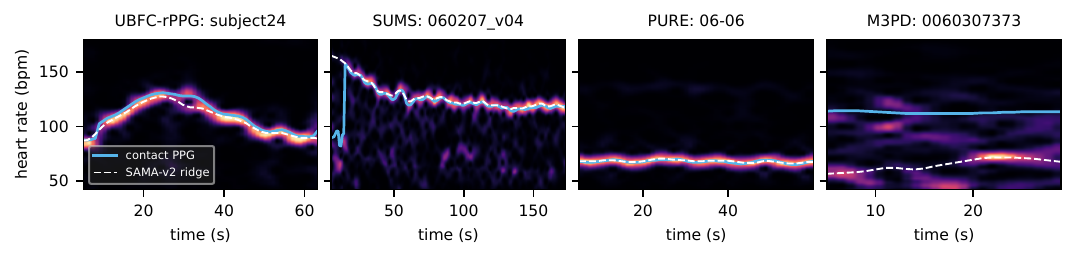}
\caption{Consensus spectrograms with the reference track (blue) and the \sama\ ridge (white dashed). Third from left: a clean PURE case. Second: the SUMS reference is wrong for the first seconds. Right: an M3PD recording in which the facial pulse is not visible and the ridge locks onto noise.}
\label{fig:ex}
\end{figure}

When heart rate drifts within a recording, the whole-video periodogram of the reference is broad and its maximum depends on the amplitude envelope of the finger sensor. To measure this we synthesised, for each recording, a constant-amplitude sinusoid that follows the reference's own 10\,s heart-rate track, and scored it against the reference with the per-video protocol. This oracle, which knows the true heart rate at every second, has an MAE of 1.25\,bpm on UBFC-rPPG, 1.16 on PURE, 1.64 on SUMS, 1.62 on the MMPD subset and 0.30 on M3PD (short clips, stable rate) (Fig.~\ref{fig:floor}). Per-video differences of the order of 1\,bpm on the first four datasets therefore reflect envelope agreement with the finger sensor as much as heart-rate accuracy, which is one reason to prefer windowed evaluation. Figure~\ref{fig:ex} shows typical tracks, including a start-up glitch in a SUMS reference and an M3PD failure.

\paragraph{Scoring band.} With the 0.6--3.3\,Hz band of the current toolbox release the per-video MAE of \sama\ is 1.92, 0.48, 1.67, 8.52 and 8.39\,bpm (UBFC-rPPG, PURE, SUMS, M3PD, MMPD subset) and that of POS is 2.73, 0.57, 1.80, 11.73 and 8.82; \sama\ is lower on all five. PURE improves for both because several of its participants have resting rates near 45\,bpm, the edge of the narrower band.

\subsection{Skin tone}
None of the five datasets as available to us supports a skin-tone analysis: only the MMPD subset carries Fitzpatrick labels, for two people. On it, the per-video MAE is 5.51\,bpm for the type~III participant (11 recordings) and 10.15 for the type~IV participant (20 recordings), a difference confounded with identity and conditions. We also computed the individual typology angle (ITA) from the mean skin colour of each recording. The values span $-90^\circ$ to $94^\circ$ and are driven by exposure and white balance (for example, the median is $-46^\circ$ for PURE, whose participants are light-skinned but dimly exposed), and their rank correlation with error is weak and inconsistent in sign ($-0.35$ to $0.12$ across datasets). Skin tone therefore has to be measured at recording time with a colour reference and a colorimeter; it cannot be recovered afterwards.

\section{Limitations}
\label{sec:lim}
(1) All results use a fixed face box; a tracked face would probably help every method and might restore the value of sub-face regions. (2) The face-wide default and AGC were adopted after seeing results on these datasets; the constants were not tuned, but the two design choices are not independent of the test data, and a pre-registered replication on new data is needed. (3) Neural baselines were not run; comparisons with them rest on published numbers under different crops and splits. (4) MMPD is a two-participant subset. (5) Errors remain above 8\,bpm on handheld phone video (M3PD, MMPD), far from clinical use without abstention. (6) The cohorts contain almost no dark-skinned participants, so nothing here shows that the method is equitable across skin tones; that is the open question our planned data collection addresses. (7) Reference pulse in all datasets is a finger oximeter, not ECG, and some reference segments are visibly wrong.

\section{Conclusion}
A chrominance pipeline with no learned parameters, extended by consensus weighting, a discount for measurable interference and ridge tracking, gives the lowest average error among classical methods on five public datasets and a usable confidence, with the clearest gains on the short windows that matter for live use. It approaches published cross-dataset networks on PURE, improves on published face-only results on M3PD, and falls short of the best networks on UBFC-rPPG and SUMS. The evidence that is still missing, and that no public dataset can supply, is accuracy across the range of Indian skin tones against ECG with calibrated colour; collecting it is the purpose of the study this work prepares.

\end{document}